\documentclass[]{spie}  
\usepackage{caption}
 
\usepackage{amsmath,amsfonts,amssymb}
\usepackage{graphicx}
\usepackage[colorlinks=true, allcolors=blue]{hyperref}

\usepackage{caption}
\usepackage{subcaption}
\usepackage[export]{adjustbox}

\usepackage{float} 

\title{A detection-task-specific deep-learning method to improve the quality of sparse-view myocardial perfusion SPECT images}

\author[a]{Zezhang Yang}
\author[b]{Zitong Yu}
\author[c]{Nuri Choi}
\author[a,b,c]{Abhinav K. Jha}
\affil[a]{Department of Electrical and Systems Engineering, Washington University, St. Louis, MO, USA}
\affil[b]{Department of Biomedical Engineering, Washington University, St. Louis, MO, USA}
\affil[c]{Mallinckrodt Institute of Radiology, Washington University, St. Louis, MO, USA}

\begin{document} 
\maketitle

\begin{abstract}

Myocardial perfusion imaging (MPI) with single-photon emission computed tomography (SPECT) is a widely used and cost-effective diagnostic tool for coronary artery disease. However, the lengthy scanning time in this imaging procedure can cause patient discomfort, motion artifacts, and potentially inaccurate diagnoses due to misalignment between the SPECT scans and the CT-scans which are acquired for attenuation compensation. Reducing projection angles is a potential way to shorten scanning time, but this can adversely impact the quality of the reconstructed images. To address this issue, we propose a detection-task-specific deep-learning method for sparse-view MPI SPECT images. This method integrates an observer loss term that penalizes the loss of anthropomorphic channel features with the goal of improving performance in perfusion defect-detection task. We observed that, on the task of detecting myocardial perfusion defects, the proposed method yielded an area under the receiver operating characteristic (ROC) curve (AUC) significantly larger than the sparse-view protocol. Further, the proposed method was observed to be able to restore the structure of the left ventricle wall, demonstrating ability to overcome sparse-sampling artifacts. Our preliminary results motivate further evaluations of the method.

\end{abstract}


\section{Introduction}
\label{sec:intro}  

Myocardial perfusion imaging (MPI) by single-photon emission computed tomography (SPECT) plays an important role in diagnosing, monitoring, and assessing coronary artery disease.\cite{zaret2010clinical} The procedure typically involves a step-and-shoot protocol with scanning time of up to 15 minutes. However, prolonged scanning presents several challenges. First, patients may experience discomfort and anxiety during the prolonged scanning procedure.\cite{nightingale2012thought} Second, the lengthy scanning procedure causes potential motion artifacts that degrade image quality.\cite{agarwal2014myocardial,botvinick1993quantitative} Additionally, in this process, often an extra CT scan is acquired for attenuation compensation. This long scanning time can result in patient motion, which can cause misregistration between the SPECT and CT scans, compromising diagnostic accuracy.\cite{goetze2007attenuation,goetze2007prevalence,saleki2019influence} Furthermore, the lengthy scanning time reduces patient throughput in high-volume clinical institutions. Hence, there is an important need for methods that can shorten the scanning time.\cite{heller2012practical}

One potential method to shorten scanning time is to reduce the number of projection angles while maintaining the same acquisition time per angle. However, the sparse-view acquisition yields reconstructed images with limited image quality for the task performance.\cite{chen2023cross} Therefore, there is an important need for methods to process sparse-view SPECT MPI images such that their quality for the task performance is improved. Recent advancements in deep-learning (DL) have shown promise in processing sparse-view images,\cite{chen2023dudoss, ryden2021deep, li2022lu} with most approaches being trained using image fidelity loss functions and evaluated through figures of merit such as root mean squared error (RMSE) and structural similarity index measure (SSIM). However, fidelity-based evaluations may not correspond to performance on clinical tasks.\cite{Yu2020,Yu2023,Li2021,Badal2019,jha2022nuclear} For instance, previous literature has found discrepancies between fidelity-based assessments of a DL-based denoising method for MPI SPECT and performance on the task of cardiac defect detection.\cite{yu2023need} Of most relevance to this study, Zhang et al. observed inconsistencies in fidelity-based evaluations versus detection-task-based evaluations of DL algorithms that processed low-resolution images to predict high-resolution samples.\cite{zhang2021impact} These findings highlight the need for investigating strategies that can incorporate the clinical task when developing DL-based imaging methods.

Recently, a detection-task-specific method for processing low-count MPI SPECT images was proposed by Rahman et al.\cite{rahman2023demist} This was a deep-learning method that utilized our understanding of the human observer performance to preserve detection-task-specific features while processing the MPI SPECT images. In a retrospective clinical study, DEMIST yielded an improvement in performance on the task of detecting myocardial perfusion defects with an anthropomorphic model observer. Furthermore, DL approaches have been proposed for CT that typically utilize hybrid loss functions, combining image fidelity terms with task-specific components.\cite{ongie2022optimizing,li2022task,han2021low} The importance of incorporating task-specific features in loss functions is supported by their demonstrated ability to enhance detection task performance, such as optimizing the signal-to-noise ratio\cite{ongie2022optimizing}, using binary cross-entropy with DL-based observers\cite{li2022task} and using a trained network to extract features and define the feature-level loss as observer loss.\cite{han2021low}

Reducing the number of scanning angles leads to the loss of high-frequency features, impairing performance on the task of defect detection.\cite{li2024heal,qu2020sparse} In this context, Li et al. developed a high-frequency enhancement method using dual-domain attention and a novel high-frequency regularization term, improving the ability of the method to recover high-frequency features under sparse-view conditions.\cite{li2024heal} Similarly, in another study, Ayyoubzadeh et al. proposed a feature accentuation space enabling image restoration CNNs to enhance sharpness and high-frequency detail clarity. This method significantly improves the visual quality of restored images, particularly along edges and high-texture regions.\cite{9583853} These studies provide promise that DL methods may help address high-frequency feature loss in imaging.

Building on prior work that considers task-specific features and the idea of the DEMIST approach in processing low-count MPI SPECT images,\cite{rahman2023demist} we propose a detection-task-specific deep-learning method to enhance myocardial perfusion defect detection task performance for sparse-view MPI SPECT images. The proposed approach incorporates a task-specific term into the loss function alongside a fidelity term. The task-specific term is inspired by previous studies that have observed that the human observer performance can be modeled as an anthropomorphic model observer that processes information through a combination of frequency-selective channels.\cite{frey2002application} The proposed method was evaluated on the clinical task of detecting myocardial perfusion defects using an anthropomorphic model observer.

\section{Methods}



\subsection{Proposed method}

\begin{figure} [ht]
   \begin{center}
   \begin{tabular}{c} 
   \includegraphics[height=5.5cm]{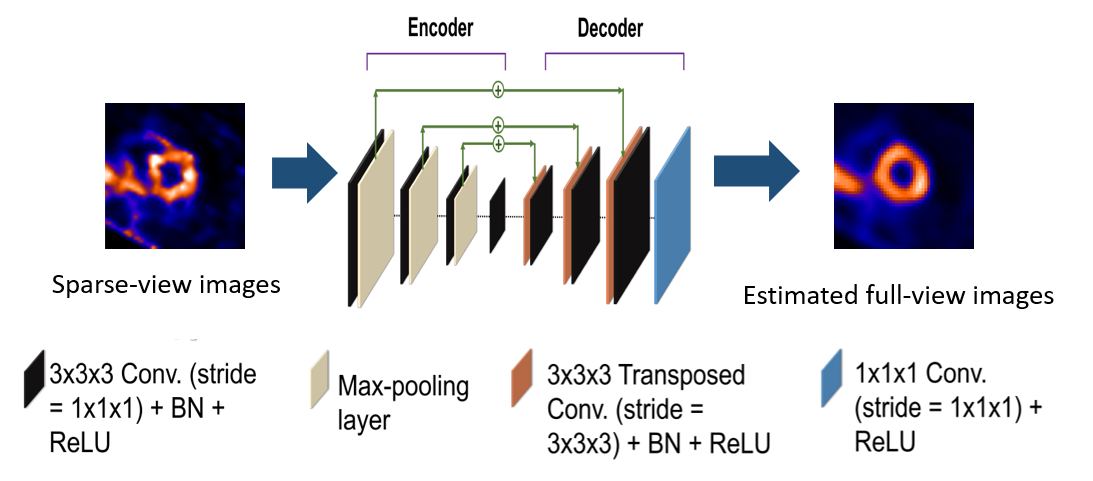}
   \end{tabular}
   \end{center}
   \caption[example] 
   { \label{framework} 
3D neural network architecture of the proposed method}
   \end{figure}

The architecture of the neural network for the proposed method is depicted in Fig.\ref{framework}. The proposed method takes sparse-view short-axis SPECT images as inputs and outputs the estimated full-view short-axis images.

The goal of the proposed method is to process the sparse-view images such that the processed image yields improved performance on the task of detecting myocardial perfusion defects. For this purpose, we employ a loss function comprising two components: a fidelity term and an observer-based loss term. The fidelity term we incorporated is the mean squared error (MSE) between the estimated full-view images and the corresponding full-view images. Psychophysical studies have shown that the human observer can be modeled as a model observer that processes images through frequency-selective channels.\cite{barrett2013foundations} The channelized Hotelling observer (CHO) with rotationally symmetric frequency channels has been observed to emulate human observer performance in MPI SPECT defect detection.\cite{myers1987addition,sankaran2002optimum} Therefore, we incorporated an observer-based loss that penalizes the MSE between the channel vectors of the estimated full-view image and the corresponding full-view image.

Denote the size of the training dataset by $J$, and the $j^{\text {th}}$ sample of the full-view image by $\hat{\boldsymbol{f}}_{FV}^j$ and the $j^{\text {th}}$ sample of the estimated full-view image by $\hat{\boldsymbol{f}}_{FV}^{est,j}$, respectively. Also, denote the channel operator by $\boldsymbol{U}$, a $C \times N_{2D}$ matrix, where $\mathrm{C}$ represents the number of channels, and $N_{2D}$ is the size of each image slice. In addition, denote the total number of slices picked from each image by $s_2$ and the shift operation at $j^{\text {th}}$ image as $S^{j}$. Finally, denote the  $j^{\text {th}}$ sample, $s^{\text{th}}$ slice of the full-view image by $\hat{\boldsymbol{f}}_{FV,2D,s}^j$ and the $j^{\text {th}}$ sample, $s^{\text{th}}$ slice of the estimated full-view image by $\hat{\boldsymbol{f}}_{FV,2D,s}^{est,j}$, respectively. Then, the loss function is given by:


\begin{equation}
    L = \frac {1}{J} \sum _ {j=1}^ {J} \left\{|| \widehat {f} _ {FV}^ {j}- \widehat {f}_ {FV}^ {est,j}||_ {2}^ {2}  \right\} +\lambda \frac {1}{J} \sum _ {j=1}^ {J} \left\{  \sum _ {s=s1}^ {s_ {2}} ||\left(\mathcal{S}^j \boldsymbol{U}\right)\left(\hat{\boldsymbol{f}}_{FV,2D,s}^j-\hat{\boldsymbol{f}}_{FV,2D,s}^{\text {est, }, j}\right)||_ {2}^ {2}\right\}.
\end{equation}

\subsection{Objective evaluation of the proposed method}
\label{Data description and preprocessing of data:}

We evaluated the proposed method in an IRB-approved retrospective study. We followed best practices for developing \cite{Bradshawjnumed.121.262567} and evaluating \cite{jha2022nuclear} the proposed method, as recommended by the Society of Nuclear Medicine and Molecular Imaging (SNMMI) AI Task Force.

We collected data from N = 449 patients, including SPECT projection and CT images. These images were originally acquired at the full-view and normal-dose level. To simulate the sparse-view images, we evenly sampled projection angles from full projection angles. These patients were divided into the training dataset of N = 184 patients and the test dataset of N = 265 patients. To ensure that the ground truth of the defects was known, we selected only projection scans of patients with no defects and inserted synthetic defects into these scans to create a population of defect-present cases. These defect types were defined based on their location in the left ventricle wall, extent, and severity. The defect insertion pipeline is described in detail in Rahman et al.\cite{rahman2023demist,narayanan2001optimization} To generate sparse-view projections, we sampled 5 projection angles evenly from an original set of 30 projection angles. The full-view and sparse-view projection data were reconstructed using the ordered subsets expectation maximization (OSEM) algorithm with 8 iterations and 5 subsets for the sparse-view projection data and 6 subsets for the full-view projection data. The reconstructions were implemented in CASToR\cite{merlin2018castor}. The resulting images were then reoriented to the short-axis view. From these reoriented images, we extracted a 48×48×48 volume centered on the left ventricle.

The neural network, as shown in Fig.\ref{framework}, was trained using the training samples. Further, five-fold cross-validation was conducted. The network was then tested with the test data. The test set comprised 134 healthy patients for the defect-absent population, while a separate set of 131 healthy patients with inserted defects was used to create the defect-present population.

Performance was evaluated on the task of detecting myocardial perfusion defects using an anthropomorphic model observer that, in previous studies, has been shown to emulate human-observer performance in MPI SPECT studies for defect detection tasks.\cite{sankaran2002optimum,wollenweber1999comparison} Performance on the detection task was quantified using the AUC values. The method was also compared with a method that only contained the MSE between the estimated and actual full-view images in the loss function and did not incorporate the observer-based loss term. We refer to this approach as the task-agnostic DL method (TADL). The statistical significance of the results was assessed using DeLong's test adjusted by Obuchowski’s method.\cite{obuchowski1997nonparametric,delong1988comparing} A $p$-value less than 0.05 was used to infer statistical significance.

\section{Results}

  \begin{figure} [ht]
   \begin{center}
   \begin{tabular}{c} 
   \includegraphics[height=9cm]{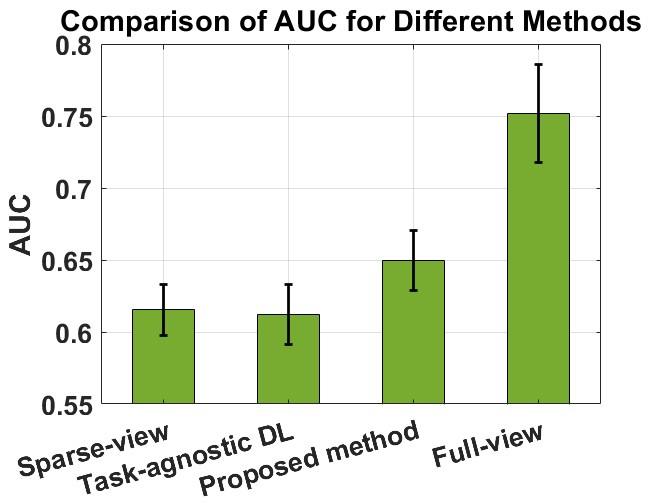}
   \end{tabular}
  \end{center}
   \caption[example] 
   { \label{fig:auc}
AUC values for the full-view and sparse-view protocol images, as well as the images processed using the proposed method and the task-agnostic DL method.}
   \end{figure}

In this section, we present results with the proposed method and compare it with other protocols at the 5 sparse-view level setting. Fig. \ref{fig:auc} shows the AUC values obtained with the sparse-view protocol, the task-agnostic DL method, and the proposed task-specific method at the 5 sparse-view level, along with the full-view protocol. The proposed method significantly outperformed the sparse-view protocol (DeLong's test $p$-value $< 0.05$).

\begin{figure*} [ht]
   \begin{center}
   \begin{tabular}{c} 
   \includegraphics[height=9cm]{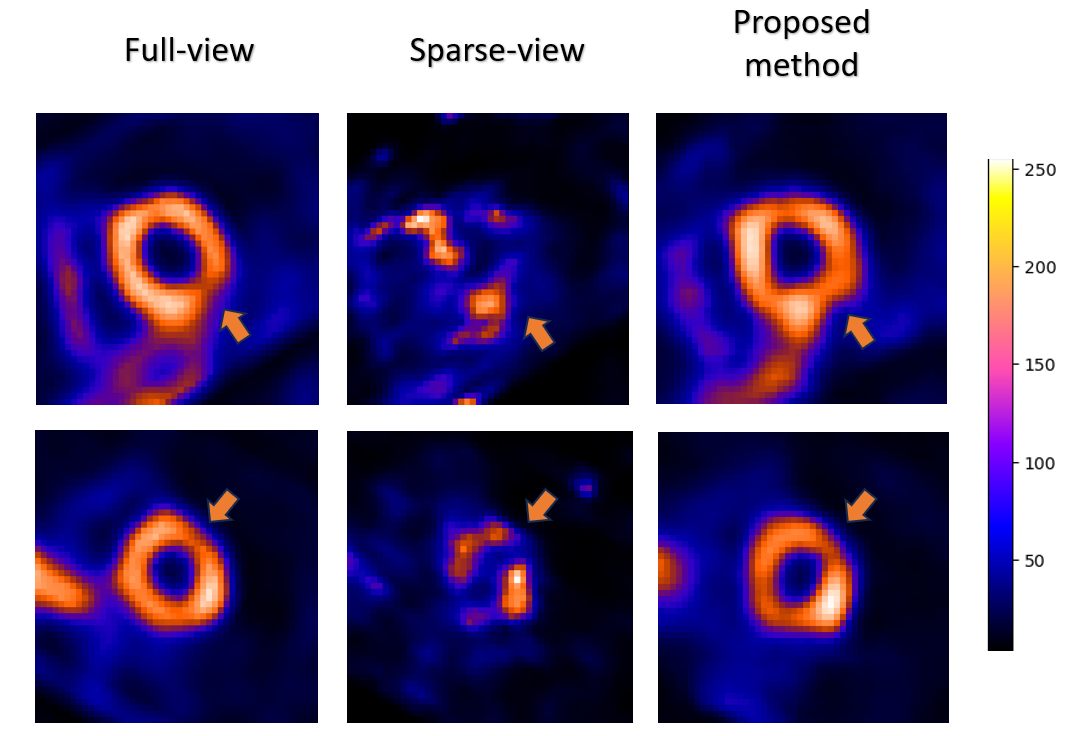}
   \end{tabular}
   \end{center}
   \caption[example] 
   { \label{fig:example}
Two representative test cases qualitatively demonstrate the performance of the proposed method. In each case, the sparse-view level was set to 5. In the uppercase and lowercase cases, defects were located in the inferior and anterior walls, respectively. For all cases, the defects had an extent of 30 degrees and a severity of 25\%.}
   \end{figure*}

Fig. \ref{fig:example} shows the qualitative comparison of the different protocols at the 5 sparse-view level setting. The proposed method demonstrates improved performance in reducing sparse-sampling artifacts. Notably, it restores the structure of the left ventricle wall while maintaining the visibility of the defect. The qualitative comparison in these two examples highlights the method's ability to enhance defect visibility compared to sparse-view images, supporting its potential to improve defect detectability.


\section{Discussion}

In this study, we developed a deep-learning-based method aimed at processing sparse-view MPI SPECT images with the goal of improving performance on the task of detecting myocardial perfusion defects. The method integrates an observer-based loss term alongside a fidelity-based term. The observer-based loss term specifically penalizes errors in features extracted from anthropomorphic channels. By minimizing these errors, the proposed method aims to improve the performance of sparse-view images on the task of detecting myocardial perfusion defects.

From the AUC results presented in Fig \ref{fig:auc}, the proposed task-specific method yields an improvement in detection task performance over sparse-view images. In particular, incorporating the observer-based loss term improves myocardial perfusion defect detection task performance from sparse-view images. From the preliminary results presented in Fig \ref{fig:example}, we see that the proposed method shows promise in improving the performance of the sparse-view images in defect detection task. From the output images of the proposed method, the structure of the overall left ventricle is restored and the defect features are still visible. Therefore, the task of defect detection is expected to be more accurate in the estimated full-view images compared to the sparse-view images. This finding visually supports the potential of the proposed task-specific method to improve the performance of processed sparse-view images on the task of myocardial perfusion defect detection.

The sparse-view protocol loses high-frequency features. The proposed method addresses this issue by using the CNN neural networks to learn contextual priors and infer missing details by estimating full-view images from sparse-full images and comparing the estimated full-view images with the actual full-view images during training. The results in Fig. \ref{fig:example} show that the proposed method has the ability to address the sparse-sampling artifacts. These observations motivate further quantitative investigation.

There are several limitations that should be considered. First, the method was trained on data with inserted synthetic defects due to challenges in getting a reliable ground truth for the presence and location of the myocardial perfusion defect in clinical images. Therefore, one area of future research is to use clinical defect-present data with well-curated ground truth. Second, the defect locations were restricted to two regions. Expanding this to include additional regions, such as the septal and lateral walls, would further support the robustness of the method's performance. Finally, the study used data from a single center. In the future, we will conduct multi-center validation as it is essential to evaluate the generalizability of the method across diverse clinical settings\cite{Zitongspie2025}.

\section{Conclusion and future work}


We proposed a task-specific deep-learning method designed to enhance sparse-view myocardial perfusion SPECT images. The loss function in the method penalizes the loss of anthropomorphic channel features with the goal of improving performance in the perfusion defect-detection task. Our preliminary findings suggest that the proposed method can improve both the quality of the sparse-view images and the accuracy of defect detection, motivating further validation of the method.

\section{ACKNOWLEDGMENTS}

This work was supported in part by National Institute of Biomedical Imaging and Bioengineering of National Institute of Health (NIH) under grant number R01-EB031051 and R01-EB031962, and National Science Foundation (NSF) CAREER Award 2239707.

\bibliography{references_modified.bib} 
\bibliographystyle{spiebib} 

\end{document}